\documentclass[journal,twoside]{IEEEtran}
\makeatletter
\long\def\@makecaption#1#2{\ifx\@captype\@IEEEtablestring%
\footnotesize\begin{center}{\normalfont\footnotesize #1}\\
{\normalfont\footnotesize\scshape #2}\end{center}%
\@IEEEtablecaptionsepspace
\else
\@IEEEfigurecaptionsepspace
\setbox\@tempboxa\hbox{\normalfont\footnotesize {#1.}~~ #2}%
\ifdim \wd\@tempboxa >\hsize%
\setbox\@tempboxa\hbox{\normalfont\footnotesize {#1.}~~ }%
\parbox[t]{\hsize}{\normalfont\footnotesize \noindent\unhbox\@tempboxa#2}%
\else
\hbox to\hsize{\normalfont\footnotesize\hfil\box\@tempboxa\hfil}\fi\fi}
\makeatother
\IEEEoverridecommandlockouts
\usepackage{amsthm, fancyhdr, amsmath,amssymb,amsfonts,bm}
\usepackage{array, booktabs}
\usepackage{tipa}
\usepackage{gensymb}
\usepackage{algorithmic}
\usepackage{graphicx}
\usepackage{multirow}
\usepackage{amsmath}
\usepackage{textcomp}
\usepackage[table]{xcolor}
\usepackage[noadjust]{cite}
\usepackage[linesnumbered,ruled]{algorithm2e}
\usepackage{hyperref}

\usepackage{dirtytalk}
\ifCLASSOPTIONcompsoc
\usepackage{caption}
\usepackage{float}
\usepackage{tabularx}
\usepackage[caption=false,font=normalsize,labelfon
t=sf,textfont=sf]{subfig}
\else
\usepackage[caption=false,font=footnotesize]{subfi
g}
\newcommand{\mat}[1]{\boldsymbol{#1}}
\fi
\newcommand{\norm}[1]{\left\lVert #1 \right\rVert}
\ifCLASSINFOpdf
 
\else
 
\fi

\begin{document}

\title{
A holistic perception system of internal and external monitoring for ground autonomous vehicles: AutoTRUST paradigm
\thanks{This work has received funding from the EU’s Horizon Europe research and innovation programme in the frame of the AutoTRUST project “Autonomous self-adaptive services for TRansformational personalized inclUsivenesS and resilience in mobility” under the Grant Agreement No 101148123.}
}

\author{
\IEEEauthorblockN{Alexandros Gkillas$^{1,2}$,  Christos Anagnostopoulos$^{1,2}$, Nikos Piperigkos$^{1,2}$, Dimitris Tsiktsiris$^{3}$,  Theofilos Christodoulou$^{3}$, Theofanis Siamatras$^{3}$, Dimitrios Triantafyllou$^{3}$, Christos Basdekis$^{3}$, Theoktisti Marinopoulou$^{3}$, Panagiotis Lepentsiotis$^{3}$, Elefterios Blitsis$^{3}$, Aggeliki Zacharaki$^{3}$, Nearchos Stylianidis$^{4}$, Leonidas Katelaris$^{4}$, Lamberto Salvan$^5$, Aris~S. Lalos$^{1}$, Christos Laoudias$^{4,5}$, Antonios Lalas$^{3}$ and Konstantinos Votis$^{3}$}\\
\IEEEauthorblockA{$^1$Industrial Systems Institute, ATHENA Research Center, Patras Science Park, Greece\\
$^2$AviSense.AI, Patras Science Park, Greece, \\ $^3$Information Technologies Institute,
CERTH, Greece\\
$^4$KIOS Research and Innovation Center of Excellence, University of Cyprus, Cyprus \\
$^5$ALKE Electric Vehicles, Padova, Italy\\
Emails: \{gkillas, anagnostopoulos, piperigkos\}@avisense.ai, lalos@athenarc.gr, \{tsiktsiris,  tchristodoulou, theosiam, dtriantafyllou, cmpasdek, tmarinop, panalepe, eblitsi, angezach, lalas,  kvotis\}@iti.gr \{stylianidis.nearchos, katelaris.leonidas, laoudias.christos\}@ucy.ac.cy, lamberto.salvan@alke.com 
}
}
\maketitle

\begin{abstract}
This paper introduces a holistic perception system for internal and external monitoring of autonomous vehicles, with the aim of demonstrating a novel AI-leveraged self-adaptive framework of advanced
vehicle technologies and solutions that optimize perception and experience on-board. Internal monitoring system relies on a multi-camera setup designed for predicting and identifying driver and occupant behavior through facial recognition, exploiting in addition a large language model as virtual assistant. Moreover, the in-cabin monitoring system includes AI-empowered smart sensors that measure
air-quality and perform thermal comfort analysis for efficient on and off-boarding. On the other hand, external monitoring system perceives the surrounding environment of vehicle, through a LiDAR-based cost-efficient semantic segmentation approach, that performs highly accurate and efficient super-resolution on low-quality raw 3D point clouds. The holistic perception framework is developed in the context of EU's Horizon Europe programm AutoTRUST, and has been integrated and deployed on a real electric vehicle provided by ALKE. Experimental validation and evaluation at the integration site of Joint Research Centre at Ispra, Italy, highlights increased performance and efficiency of the modular blocks of the proposed perception architecture.

\textbf{Keywords:} Autonomous vehicles, Perception, Multi-camera, LiDAR, Virtual assistant
\end{abstract}

\section{Introduction}

Autonomous vehicles (AVs) are rapidly transforming modern societies, driven by their potential to significantly enhance road safety, improve traffic efficiency, and provide greater accessibility and mobility options for diverse populations. With urban and transportation conditions becoming more complex every, the adoption of AV technologies has the potential to mitigate traffic congestion, reduce environmental impact, and contribute to more sustainable urban development \cite{8951131, 9046805}. In this context, effective and reliable perception systems are critical components of AVs, ensuring safe navigation and robust interaction with complex driving environments. Internal monitoring systems play a vital role in tracking driver behavior, occupant health, and comfort, thereby enhancing safety and personalized user experiences within autonomous or semi-autonomous vehicles. These systems typically utilize AI-powered multi-camera setups, facial recognition, emotion analysis, and environmental sensors to proactively respond to in-cabin dynamics \cite{8751968, 8678436}. Furthermore, external monitoring systems, such as LiDAR and camera-based perception approaches, are essential for accurate scene understanding, obstacle detection, and precise navigation decisions in dynamic and clattered external environments \cite{8793495}.

Recent advancements in internal and external vehicle perception have significantly enhanced autonomous driving capabilities, yet critical challenges remain in real-time processing, integration feasibility, and deployment on embedded platforms. Lightweight CNNs and transformer-based models have enabled in-cabin activity, emotion, and identity recognition under resource constraints \cite{fer3, fid4}, while efficient sound event detection supports driver safety through low-latency audio analysis \cite{sed1, sed2}. Additionally, intelligent in-cabin environmental monitoring which is critical for safety and passenger comfort, still poses several challenges related to sensor calibration drift, space and power limitations, and integration complexity \cite{diviacco2022monitoring}. Regarding the perception of external environment, super-resolution techniques are prominent for enhancing semantic segmentation from sparse low-cost LiDAR data \cite{myTIV, SHAN2020103647}, yet their adoption in embedded AV systems is still limited. Parallel research on virtual assistants has explored multimodal, emotionally aware systems \cite{kepuska2018next, duguleanua2020virtual}, with recent efforts focusing on secure, user-centric interaction models for enhanced engagement and well-being \cite{simon2023voice, bokolojnr2024user}. However, their application on AVs is still not explore enough. As such, this paper introduces a unified and deployable framework that integrates in-cabin affect sensing, virtual assistance, environmental monitoring, and efficient LiDAR super-resolution, all validated on a real electric vehicle platform. Therefore, the main contributions can be summarized as:

\begin{itemize}
    \item A holistic and unified framework for internal and external perception is introduced, suitable for diverse autonomous driving applications. 
    \item An advanced internal perception framework that combines multi-camera based emotion recognition, AI-driven virtual assistance, and intelligent air-quality sensing, significantly enhancing occupant comfort, personalization, and in-cabin experience.
    \item A novel external perception system employing a highly efficient super-resolution approach to significantly improve semantic segmentation accuracy from low-cost LiDAR sensors, enhancing autonomous vehicle navigation and environmental awareness.
    \item The full multi-modal sensor suite, including a 16-channel Velodyne LiDAR, multi-camera system and environmental sensor, has been integrated and deployed in the real electric vehicle of ALKE. Evaluation and demonstration results indicate very promising performance in terms of emotion recognition, air and thermal quality, as well as accurate semantic segmentation using low-cost LiDAR. 
\end{itemize}

\section{AutoTRUST system architecture }
\label{system_arch}

During the EUCAD 2025\footnote{\url{https://www.connectedautomateddriving.eu/}}, the first integrated implementation of the AutoTRUST project was presented on a test vehicle. The demo showcased the key architectural elements deigned to enhance trust, safety and user engagement in mobility systems. The setup aimed to demonstrate how real-time perception, adaptive reasoning and explainable interaction can be achieved on embedded hardware using a combination of multi-modal sensing, edge computing and a Large Language model. 

The test vehicle was equipped with four types of sensors. For the external monitoring a LiDAR was used providing 3D point cloud data for 360 situational awareness. Internally the cabin was monitored by two RGB cameras for face identification, emotion recognition and driver distraction modules, one microphone array for the abnormal sound event detection and an environmental sensor for the air quality monitoring. All sensor streams were processed in real time using modular perception pipelines locally into an NVIDIA Jetson Orin AGX, a high-performance edge computing unit. This setup allowed the system to operate locally without reliance on  cloud infrastructure and thus addressing concerns e.g., latency, privacy and service continuity in low connectivity environments.  

\begin{figure}
    \centering
    \includegraphics[scale=0.2]{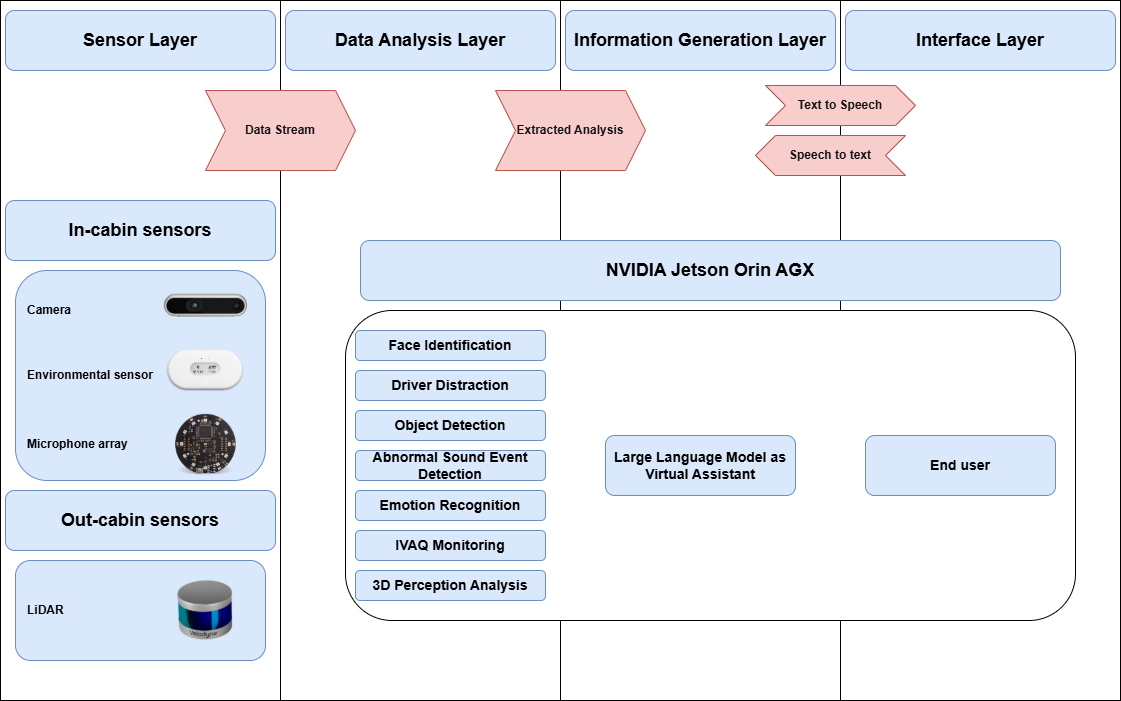}
    \caption{%
       AutoTRUST Project Architecture%
    }
    \label{fig:Autotrust_architecture}
\end{figure}

A key feature of the system’s architecture is the integration of a Large Language model, a quantized version of Llama 3 8b, hosted on the Jetson Orin unit. This Large Language Model acts as a Virtual Assistant and serves as the interface between the system and the users by providing context-aware, natural languages responses based on real-time data from the perception modules. This implementation, presented in Figure~\ref{fig:Autotrust_architecture} focuses on demonstrating the feasibility of intelligent, transparent, and human-centric automated mobility systems running entirely at the edge. It also serves as a baseline architecture for further development and evaluation of the AutoTRUST project. 

\section{Internal based perception system }
\label{internal}

\subsection{In-vehicle occupants behavior monitoring}
A custom in-cabin RGB dataset was recorded using a dashboard-mounted Intel RealSense camera, capturing a consistent side-to-mid view of the driver. It includes clips from multiple drivers, vehicles, and daylight conditions, segmented into 3-second (72-frame) sequences spanning six distraction classes: normal driving, texting, phone call, drinking, smoking, and reaching. Each frame is annotated to support temporal supervision. Preprocessing included resizing, normalization, and moderate spatial augmentations. A lightweight MoViNet-A3 model, was fine-tuned by updating only its final layers. It achieved 89\% clip-level accuracy on a held-out test set, with per-class F1 scores within ±5\% of the mean, confirming strong generalization and real-time suitability.

For real-time affect analysis, a ResEmoteNet-based facial emotion recognizer was used. This CNN architecture includes 3×3 convolutions with squeeze-and-excitation gating, residual connections, adaptive pooling, and a softmax classifier. The system detects seven universal expressions: happy, surprise, sad, anger, disgust, fear, and neutral, from the same in-cabin camera feed. Without task-specific fine-tuning, the model reached 72.9\% accuracy on AffectNet-7 and was exported to ONNX format for deployment on embedded devices.

Real-time occupant identification is handled entirely on-device. Faces are detected via an MTCNN pipeline optimized for Jetson platforms, followed by cropping and normalization. The processed images (160×160) are passed through an ONNX-converted FaceNet model (Inception-ResNet v1), producing 512-dimensional embeddings. For each user, prototype embeddings are created by averaging samples from varied poses and lighting. Inference compares incoming embeddings against stored prototypes using cosine similarity with a 0.55 threshold. The pipeline consistently achieves 96.5\% recognition accuracy under diverse conditions, with sub-50ms latency per identity using CUDA acceleration.

A YOLOv8-s model \cite{obj1, obj2}, comprising approximately 11M parameters, is used to detect both passengers and common items (e.g., phones, backpacks, laptops). Pretrained on COCO, the model is quantized to INT8 using TensorRT and a short calibration phase on domain-specific data. The full pipeline runs on a Jetson-class device with latency under 15ms and power below 15W. Internal evaluation  yields a mean AP of 92.3\% at IoU 0.5. Model parameters are managed via a single YAML file to ease field updates.

The sound event module classifies eight target classes—Baby cry, Noise, Scream, Siren, Snoring, Speech, Traffic, and Horn—selected for their relevance to driver awareness and safety \cite{sedcomp1}. Data from open repositories \cite{sedcomp2, sedcomp3, sedcomp4, sedcomp5, sedcomp6, sedcomp7, sedcomp8, sedcomp9, sedcomp10, sedcomp11} were used, totaling 3h22m of labeled audio. Clips were normalized, converted to 16kHz mono WAV, and trimmed to 5s. The pipeline uses YAMNet \cite{sedcomp12} to extract 1024D embeddings from waveforms, which are classified by a four-layer MLP with ReLU, batch normalization, and dropout. Trained with stratified splits (70/20/10) using Adam optimizer, the model showed stable convergence and strong generalization in noisy in-cabin conditions.

\subsection{In-vehicle air quality and thermal conditions monitoring}
The internal-based IVAQ monitoring and reporting system developed  is composed of two core components: the Connecting Space Hub and a distributed network of smart environmental sensors. The smart sensors are responsible for collecting high-resolution data on key parameters that influence passenger comfort and safety, including temperature, CO2, PM2.5, PM1, VOCs, air pressure, ambient noise, light intensity, and radiation. These sensors perform live tracking with a minimum update frequency of five minutes and communicate their measurements to the Connecting Space Hub. The Hub acts as the central processing and communication unit, managing data flow, enabling secure API-based export of both live and historical datasets, and supporting seamless integration with external visualization platforms. This architecture ensures that real-time environmental data is always accessible for monitoring, analysis, and decision-making.
The system provides several key outcomes essential for optimizing in-cabin environments. First, it enables detailed analysis of thermal comfort and air quality trends across different cabin zones and time intervals. By continuously evaluating spatial and temporal data, it becomes possible to identify specific areas of contamination or discomfort—such as hotspots of poor air quality or temperature imbalance. Furthermore, the system supports the formulation of targeted strategies to improve environmental conditions, enhancing both passenger safety and ride quality. Finally, the correlation analysis capabilities allow operators to explore how environmental variables relate to factors like time of day, vehicle occupancy levels, and seating distribution, offering valuable insights for adaptive ventilation control, dynamic seating arrangements, and overall passenger experience optimization.
To validate the system under real-world conditions, it was deployed on a real vehicle.
Real-time environmental data was successfully collected and processed during operation, highlighting the system’s capacity to inform data-driven in-cabin climate management. For visualization purposes, a dedicated Grafana interface was developed. This dashboard not only presented real-time measurements but also computed and displayed hourly averaged values to support trend analysis and retrospective evaluation. The interface offered intuitive, user-friendly access to both live and historical data, facilitating continuous environmental monitoring. 

\section{Virtual Assistant}
\label{assistant}
\subsection{Overview of Functionality }
An intelligent Virtual Assistant was developed, designed to support real-time interaction with the driver by monitoring emotional state, potential distractions, and ambient sounds. The core of the system is powered by a large language model (Meta-Llama-3-8B-Instruct \cite{grattafiori2024llama}), adapted for efficient execution on the NVIDIA Jetson platform through quantization with NanoLLM \cite{nano_llm_2024} and TVM-based compilation via MLC \cite{mlc_llm_2024}. The assistant supports seamless spoken communication by combining NVIDIA Riva \cite{nvidia_riva_2024} for speech recognition and Piper \cite{piper_nodate} for natural-sounding speech synthesis. All components are containerized using Docker, and audio input/output and system communication are managed via a Python web interface. 

\subsection{Multimodal Input and Context-Aware Reactions }

Numerous sources, including a microphone for sound classification (such as horn), a driver-facing camera for activity monitoring, and a facial emotion recognition module, provide the assistant with continuous input. A majority-vote method is applied to each of the input streams in order to smoothen any transient noise while ensuring stability and robustness. Whenever the system detects major behavioral-emotional-ambient-audio change, it will create a short context-based prompt that will be fed into the language model. The final response is intended to be succinct, educational, and conversational. 

\subsection{Efficient Language Model Deployment with NanoLLM }

Deploying a large language model on edge hardware requires careful optimization. Thus, we used the NanoLLM framework \cite{nano_llm_2024}, which provides a unified interface over various inference backends. Using int4 quantization with MLC \cite{mlc_llm_2024}, the Llama-3 model \cite{grattafiori2024llama} was compressed into a runtime-optimized format through TVM, resulting in major reductions in memory and computation demand. This methodology allows producing quality generated responses while ensuring the fast inference time on Jetson. It features a modular architecture such that ease of migration to different hardware configurations is possible with dynamic backend switching.

\subsection{NVIDIA Riva Speech-to-Text }

NVIDIA Riva \cite{nvidia_riva_2024} was used for speech recognition in order to provide quick and precise transcription. Real-time audio processing via a gRPC streaming client that connects to a local server housed in a Docker container is made possible by Riva's GPU-accelerated architecture. The system achieves high transcription accuracy with minimal delay by utilizing ASR models that are optimized for Jetson. Riva is a viable option for embedded, latency-sensitive applications due to its fully offline capability and effective GPU utilization. 

\subsection{Text-to-Speech with Piper }

To vocalize the assistant’s responses, we integrated Piper \cite{piper_nodate}, a neural text-to-speech system optimized for edge devices. Piper employs phoneme-to-waveform synthesis using grapheme-to-phoneme models and high-fidelity neural vocoders. The voice model used was precompiled into a minimal runtime representation, enabling low-latency inference directly on ARM-based Jetson devices without relying on external services. The TTS engine is tightly integrated with the assistant’s response pipeline, allowing natural and responsive audio playback. This ensures consistent performance in fully offline environments, making Piper well-suited for real-time interaction in automotive contexts. 

\section{External perception system based on LiDAR }
\label{external}
The external perception system in AutoTRUST is centered around a model-based LiDAR super-resolution network, designed for real-time and resource-efficient operation with low-resolution sensors. LiDAR sensor is employed for performing semantic segmentation, a fundamental task in 3D perception, where each point in a LiDAR scan is assigned a class label such as road, vehicle, pedestrian, or vegetation \cite{jhaldiyal2023semantic}. This process transforms unstructured point cloud data into a structured and interpretable representation of the environment, which is critical for safe navigation and decision-making. The effectiveness of segmentation models is highly dependent on the quality and density of the input data. Sparse LiDAR scans can result in missed detections and reduced accuracy, particularly for objects that are small in the scene \cite{jhaldiyal2023semantic}.

At its core, the super-resolution (SR) module formulates the enhancement of LiDAR data as an optimization problem in the range-view domain. Specifically, the relationship between the high-resolution range image $T \in \mathbb{R}^{64 \times N}$ (from a 64-channel LiDAR) and the low-resolution input $S \in \mathbb{R}^{16 \times N}$ (from a 16-channel LiDAR) is modeled as:
\begin{equation}
    S = D T + E,
\end{equation}
where $D$ is a downsampling operator and $E$ captures noise or modeling errors. The SR task is cast as the following regularized optimization problem \cite{myicip}:
\begin{equation}
\underset{\mat{T}}{\arg\min}\,\,\, \frac{1}{2} \norm{\mat{S} - \mat{D} \mat{T}}_F^2  + \mu \mathcal{J}(\mat{T}), 
\label{eq:main_problem_seg}
\end{equation}
where the first term enforces data consistency and $J(T)$ is a regularizer encoding intrinsic properties of high-resolution range images.  To solve this problem efficiently, a deep unrolling framework is adopted, where each layer of the SR network mimics one iteration of an optimization solver. The update rules at iteration $k$ involve two main modules:
\begin{align}
    T^{(k+1)} &= (D^\top D + b I)^{-1}\left(D^\top S + b Z^{(k)}\right), \\
    Z^{(k+1)} &= G_\theta\left(T^{(k+1)}\right)
\end{align}
where $G_\theta$ is learnable neural networks for denoising.

The resulting high-resolution range image is then processed by a lightweight semantic segmentation network tailored for efficiency and real-time inference. Specifically, the segmentation backbone adopts a projection-based architecture, such as LENet \cite{zhao2021fidnetlidarpointcloud} , which is designed to operate directly on 2D range images derived from LiDAR data. The encoder employs Multi-Scale Convolution Attention (MSCA) modules, combining depthwise and strip convolutions to capture both local and multi-scale contextual information. The decoder utilizes an Interpolation and Convolution (IAC) strategy, which progressively upsamples feature maps using bilinear interpolation and efficiently merges contextual cues through convolutional layers. This structure enables the network to preserve spatial resolution and semantic detail with a minimal computational footprint, making it highly suitable for embedded platforms in autonomous vehicles. 
All modules are developed as modular ROS2 nodes, supporting efficient integration with the vehicle’s broader perception stack. 
Overall, the architecture combines efficient mathematical modeling, deep learning, and modular robotics software to deliver accurate, real-time 3D perception from affordable LiDAR sensors.

\section{Experimental validation and EVALUATION}
\label{results}
\begin{figure}
    \centering
    \includegraphics[scale=0.18]{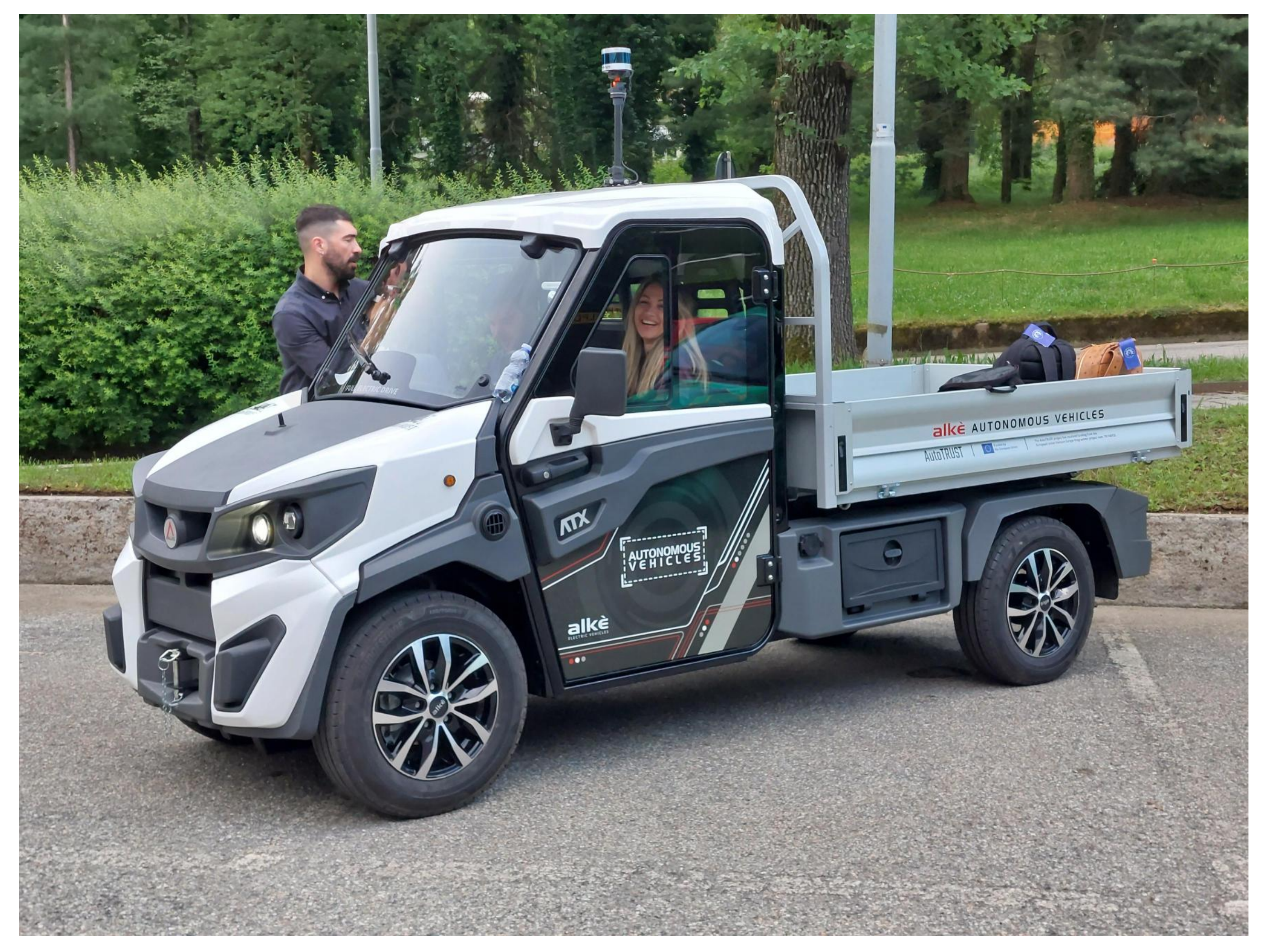}
    \caption{%
        ALKE's autonomous vehicle ATX440 EV, employed for deploying internal and external perception during EUCAD.
    }
    \label{fig:alke_vehicle}
\end{figure}

The proposed holistic perception framework has been integrated and deployed on ALKE's electic vehicle (EV) ATX440 UGV (shown in Fig.~\ref{fig:alke_vehicle}). The latte is is based on the new ATX4 range of electric vehicles launched on the market in 2025. It has a specific set-up based on the development of the AutoTRUST project. This EV platform is equipped with sensors inside and outside the vehicle to increase detection capabilities, occupant safety and accessibility, and is designed to integrate automated functionalities.
The ATX4 vehicle range is designed for highly flexible applications, from urban mobility to industrial operations and support in critical environments in remote geographical areas.

For this internal based perception system, a custom multi-driver dataset was used to train the distraction detection module under diverse lighting and driving conditions. The model achieved 89\% clip-level accuracy across six distraction classes, with per-class F1 scores within ±5\% of the mean and consistent performance even in underrepresented behaviors like smoking or reaching. Temporal robustness was validated on unseen sequences, confirming reliable real-time detection. Facial emotion recognition was benchmarked on AffectNet-7 using the ResEmoteNet model, reaching 72.9\% accuracy without domain-specific tuning. Deployed on embedded hardware, it delivered consistent low-latency inference (~30ms/frame) in real driving scenarios, effectively identifying emotions such as surprise and sadness. Occupant face recognition was tested with pre-enrolled participants under varied lighting and poses. The system achieved 96.5\% accuracy at a 0.55 cosine similarity threshold, maintaining low false acceptance/rejection rates. Running on Jetson Orin, the full pipeline executed under 50ms per identity, confirming real-time readiness. Object detection achieved 92.3\% mAP@0.5 on a proprietary in-cabin dataset. Running entirely in TensorRT with optimized pre- and post-processing, the system maintained <15ms latency at 15W, offering a zero-training solution with embedded YOLO-level accuracy. Sound event detection achieved 94\% macro-average accuracy over eight classes on an unseen 20-minute test set. Most classes exceeded F1 = 0.90, with a slight dip for “Scream” due to overlap with other acoustic events. The ONNX-runtime deployment on Jetson Orin ensured low-latency, continuous monitoring performance. The Virtual Assistant was validated through live demos and the EUCAD exhibition, providing real-time insights into distraction, emotion, and sound anomalies. 

\begin{figure}
    \centering
    \includegraphics[scale=0.18]{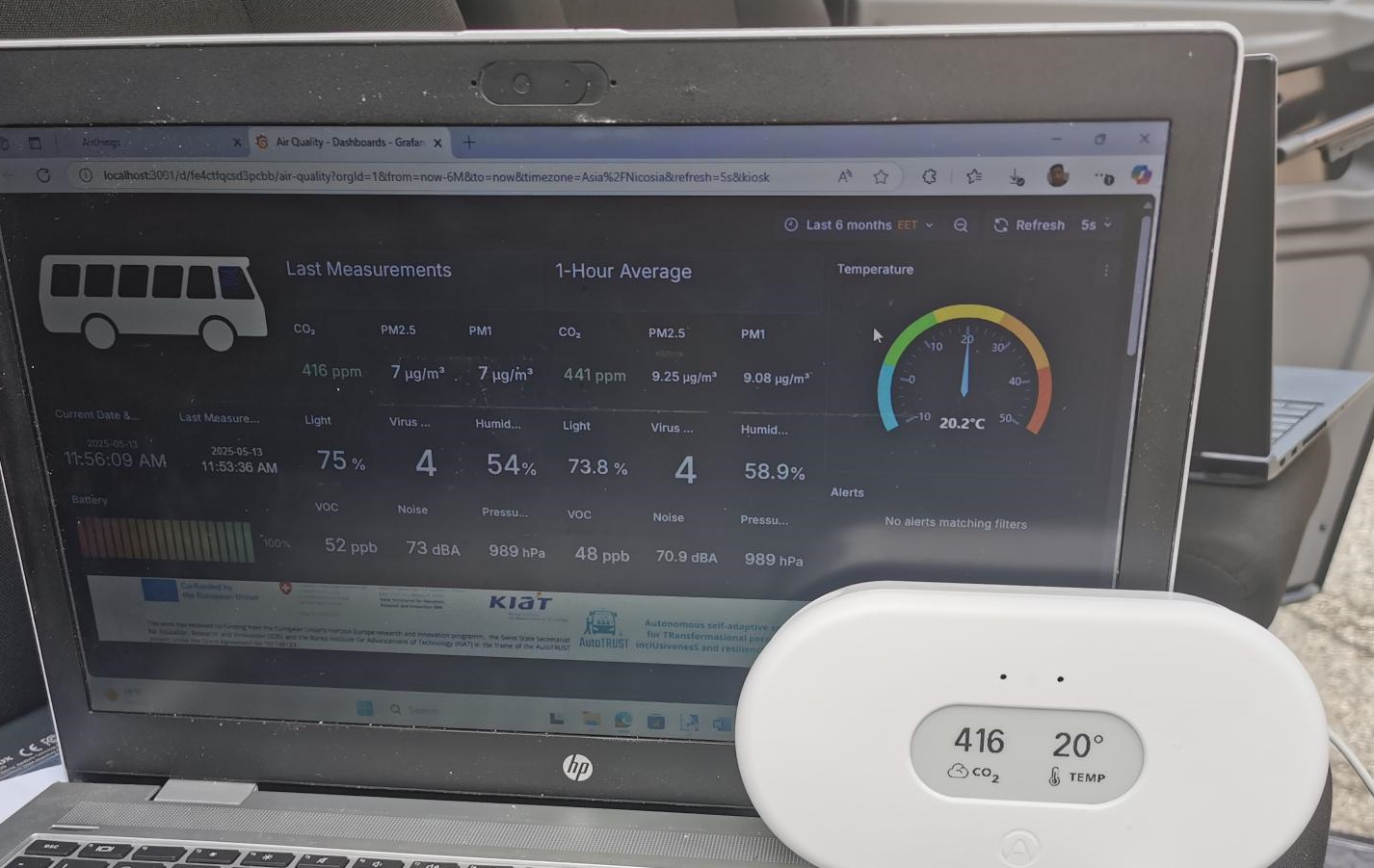} 
    \caption{IVAQ smart sensor and monitoring interface.}
    \label{fig:eucad_kios}
\end{figure}

The IVAQ system was evaluated using high-resolution in-vehicle data, including temperature, $CO_2$, PM2.5, and VOCs. From this, indices such as the Thermal Comfort Index were computed to quantify deviation from optimal cabin conditions. Threshold violations, such as elevated $CO_2$ or VOC concentrations, triggered real-time alerts for corrective action. The system enabled both immediate occupant feedback and long-term analysis, incorporating contextual factors like weather and urban exposure to inform adaptive ventilation strategies. Its embedded implementation ensured continuous monitoring with minimal overhead, contributing to occupant safety, health, and comfort in dynamic driving environments. The sensor used and the corresponding visualization interface implemented for this test is illustrated in Fig.~\ref{fig:eucad_kios}

The external perception pipeline is implemented using a dockerized ROS2-based architecture, which ensures modularity, portability, and maintainability. The data acquisition process begins with a ROS2 node that interfaces directly with the LiDAR sensor and publishes raw point clouds in real time. This point cloud data is then forwarded to a super-resolution module, implemented as a separate ROS2 node, which employs a deep learning model trained to upsample the 16-channel point clouds to a virtual 64-channel representation. Following the super-resolution step, the upsampled point clouds are processed by a 3D semantic segmentation network. This network, encapsulated in another ROS2 node, assigns semantic labels to each point in the cloud, distinguishing between key elements such as roads, vehicles, pedestrians, and vegetation. The inference is executed efficiently on the Jetson device installed inside ALKE's vehicle, enabling real-time operation within the vehicle.
To support remote monitoring and intuitive visualization, the final semantically segmented point cloud is published over a WebSocket using the rosbridge server. The data stream is visualized on an external computer using Foxglove Studio \cite{FoxgloveVisualizationObservability}, which is a ROS-compatible visualization tool that provides a powerful interface for developers and operators to inspect real-time 3D segmentation outputs, verify sensor alignment, and debug the deployed modules.

\begin{figure}
    \centering
    \includegraphics[scale=0.34]{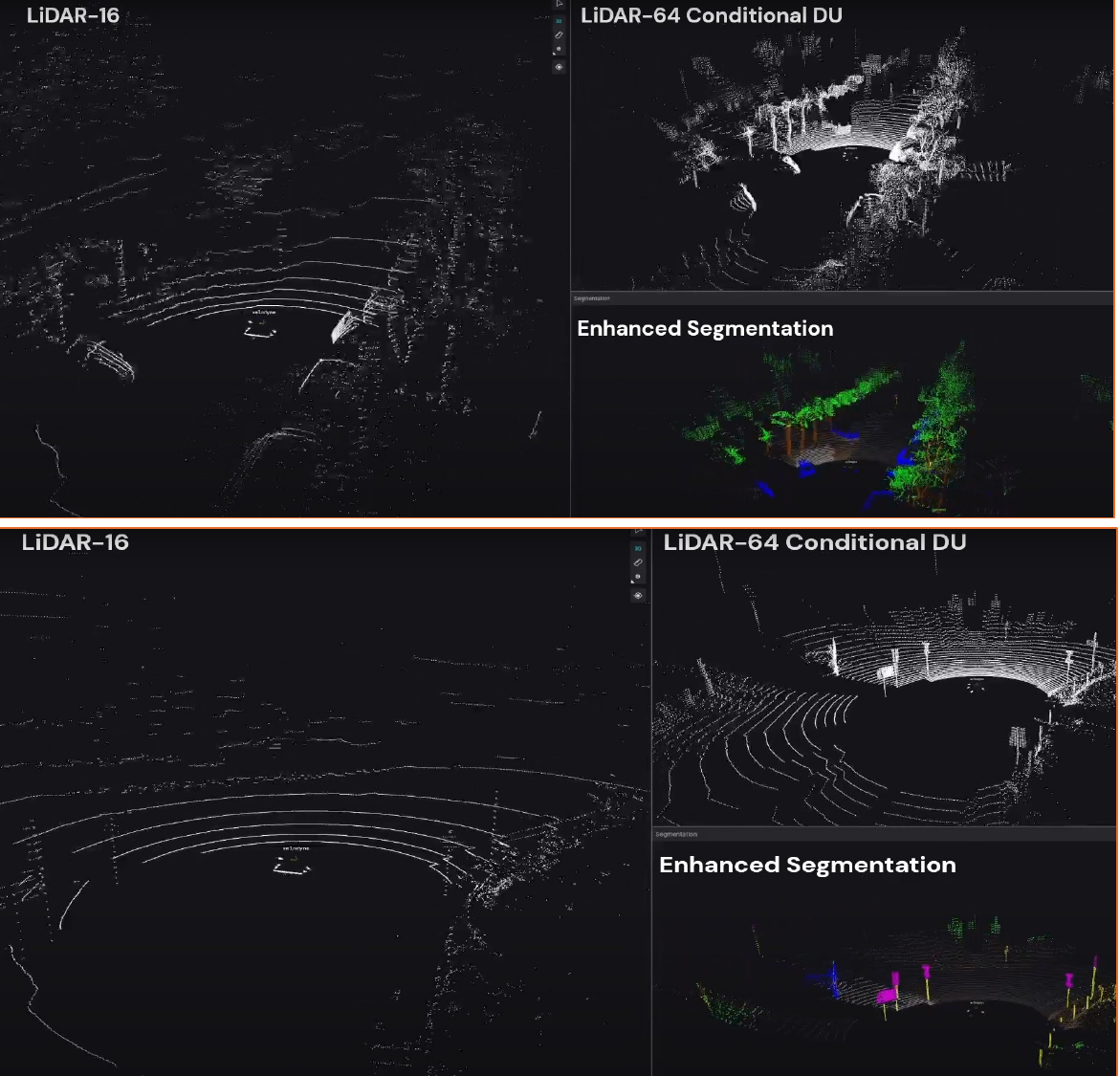}
    \caption{%
        Real-world deployment of the proposed end-to-end framework on the ALKE autonomous vehicle. Top: raw input from a 16-channel LiDAR sensor. Bottom: super-resolved output and corresponding semantic segmentation.%
    }
    \label{fig:demo}
\end{figure}

To validate the practical effectiveness of the proposed super-resolution and segmentation framework, we deployed our method on an autonomous vehicle platform provided by ALKE. The vehicle is equipped with a low-resolution, 16-channel LiDAR sensor and an embedded NVIDIA Orin computing unit, reflecting real-world constraints typical of cost-effective autonomous mobility solutions. As illustrated in Figure~\ref{fig:demo}, our approach consistently yields segmentation outputs that are both more stable and semantically coherent.  Importantly, our method achieves real-time operation at approximately 8--9 frames per second (FPS) on the embedded hardware. As further illustrated in Figure~\ref{fig:demo}, a live demonstration video of our method deployed on the ALKE autonomous vehicle is also available: \url{https://www.youtube.com/watch?v=zt9rJj7tlPQ}.

\section{Conclusion}
\label{conclusion}
This paper introduced a comprehensive holistic perception framework within the AutoTRUST project, effectively integrating advanced internal and external monitoring systems for autonomous vehicles. The internal monitoring system leverages AI-driven multi-camera setups for emotion recognition, driver distraction detection, occupant identification, and smart sensors for air quality and thermal comfort analysis, significantly enhancing passenger safety, comfort, and personalized experiences. In addition, the external perception system employs an innovative LiDAR-based super-resolution technique, addressing the limitations of low-resolution sensors by producing high-resolution semantic segmentation capabilities. The entire perception system incorporated a virtual assistant with the aim of real-time supporting the humans inside the vehicle. Real-world demonstrations in ALKE's vehicle demonstrated the framework's capabilities in terms of accuracy, real-time performance, robustness, and the potential for increased efficiency in diverse traffic and urban environments.

\bibliographystyle{IEEEtran}                
\bibliography{Bibliography}    

\end{document}